# Reinforcement Learning for Navigation of Mobile Robot with LiDAR

Inhwan Kim
*The Cho Chun Shik Graduate School of Green Transportation*
*Korea Advanced Institute of Technology*
Daejeon, South Korea
inhwan_kim@kaist.ac.kr

Sarvar Hussain Nengroo
*The Cho Chun Shik Graduate School of Green Transportation*
*Korea Advanced Institute of Technology*
Daejeon, South Korea
sarvar@kaist.ac.kr

Dongsoo Har
*The Cho Chun Shik Graduate School of Green Transportation*
*Korea Advanced Institute of Technology*
Daejeon, South Korea
dshar@kaist.ac.kr

*Abstract*—**This paper presents a technique for navigation of mobile robot with Deep Q-Network (DQN) combined with Gated Recurrent Unit (GRU). The DQN integrated with the GRU allows action skipping for improved navigation performance. This technique aims at efficient navigation of mobile robot such as autonomous parking robot. Framework for reinforcement learning can be applied to the DQN combined with the GRU in a real environment, which can be modeled by the Partially Observable Markov Decision Process (POMDP). By allowing action skipping, the ability of the DQN combined with the GRU in learning key-action can be improved. The proposed algorithm is applied to explore the feasibility of solution in real environment by the ROS-Gazebo simulator, and the simulation results show that the proposed algorithm achieves improved performance in navigation and collision avoidance as compared to the results obtained by DQN alone and DQN combined with GRU without allowing action skipping.**

*Keywords— Reinforcement learning, Deep Q-Network, Gated Recurrent Unit, Action skipping, Navigation, Path planning*

## I. Introduction

Robots are being used in various fields of our daily life, like cleaning, rescuing, medical care, delivery, autonomous driving, etc [1-4]. Particularly, in the case of mobile robots, the ability of robots to autonomously navigate in an unknown environment is essential for emerging applications such as autonomous parking by a mobile robot. Advances in sensors like LiDAR and the evolution of computer vision technologies have brought progresses to the autonomous driving of vehicles and autonomous navigation of robots [5]. Particularly, the LiDAR sensor is an optical scan sensor that measures the distances to the targets surrounding it by using radiated light. It allows high accuracy in measuring the distances to the targets, so often used for Simultaneous Localization and Mapping(SLAM) of mobile robots in real environments [6-8].

However, there are still many challenges to autonomous driving and navigation in the real environment. The traditional navigation method consists of localization, map building, and path planning. Unlike navigation with satellite systems typically associated with low transmission rate and multipath, which can be addressed by OFDM signal transmission [9,10] as well as anti-multipath technique, terrestrial navigation of mobile robot is more concerned with path planning.

One of the solutions to address efficient autonomous navigation is Reinforcement Learning(RL). The RL is one of the Deep Learning(DL) techniques and these techniques have been applied to various applications such as communication resource management, power management, and robotics [11]. By this technique, mobile robots can attempt end-to-end navigation while achieving collision avoidance, motion planning to the destination, and path planning. Path planning to destination often requires optimization in terms of total energy consumption and/or accumulated path length. Algorithms suitable for optimal path planning for various applications have been discussed in [12-15].

The RL is the technique of training the agents to make continuous decisions through trial and error in dynamic environments. Agents learn to achieve the designated goals in an uncertain environment through reward assignments to actions of the agents. Fig. 1 shows the basic structure of RL. The agent selects an action based on the current state, which is the information describing the environment.

To date, many algorithms have appeared to solve RL problems, including Q-learning, and policy gradient methods [16, 17]. These algorithms are analyzed in linear function approximation [16]. However, real-world problems are often too

This work was supported by the Institute for Information communications Technology Promotion (IITP) grant funded by the Korean government (MSIT) (No.2020-0-00440, Development of artificial intelligence technology that continuously improves itself as the situation changes in the real world).

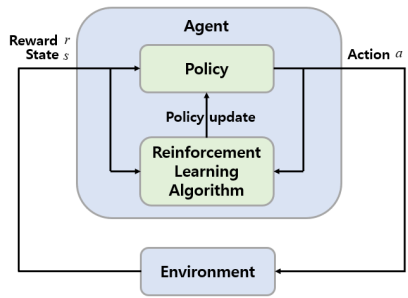

Fig. 1. Basic structure of Reinforcement learning.

complicated to be solved by linear approximation because of the high-dimensional inputs.

Of the RL algorithms, Deep Q Learning using Deep Q Network(DQN) can successfully solve the approximation of the nonlinear function [19, 20]. The DQN increases the training stability by decomposing the RL problem into sequential supervised learning. The DQN is an RL algorithm that approximates the state-action value Q through DL [18].

In Q-learning, all values according to the relation between state and action are updated and stored in a table(Q-Table). The policy is configured in a way that the action is selected by referring to the Q-table. It is a method of updating the Q-Table to maximize the cumulative reward while proceeding with the episode. The DQN, instead of updating the Q-Table, is trained to approximate the Q value. By using the DQN, high-dimensional data such as images can be used as input. The DQN trained by using the 210 x 160 pixel image input in the Atari 2600 game performed well without preference to any specific architecture or hyperparameter adjustment [20].
However, in the real environment, the agent does not have full state information, and thus has only limited observations. Therefore, the Deep Recurrent Q-Network(DRQN) method that learns only with limited information observed by the agent, which is defined as Partially Observable Markov Decision Process(POMDP) has been proposed [21]. The DRQN is in fact obtained from the combination of the Long Short-Term Memory(LSTM) network and DQN to used in the environments where the concept of the POMDP holds [21].

To deal with the long term dependency of the Recurrent Neural Network(RNN), introduced in 1997 [22], the LSTM is regarded as a variant improving the performance of the RNN. The LSTM network converges quickly in training and detects long-term dependencies in the data [23]. The LSTM has created a structure called "cell state" to resolve long-term dependencies. The cell, a unit consisting of the LSTM network, contains three gates; input, forget, and output gates. The gates store each state value in a memory cell and adjust the coefficients in contact with data to reduce unnecessary operations and errors, thereby reducing long-term dependency.

The GRU is a revised model of the LSTM network[24]. Since it has only reset gate and update gate, it has fewer parameters than the LSTM network, but shows better performance with small datasets [25,26].

The main contribution of this paper can be summarized as follows.

1. This work introduces concept of reinforcement learning applied to autonomous parking robot. To date, few works have dealt with the operation of autonomous parking robot.

2. Path planning algorithm for a single robot is presented using the DQN combined with GRU allowing action skipping.

3. Pseudo-code of the DQN combined with GRU allowing action skipping is presented to describe the technical details of the proposed algorithm.

4. Performance of the proposed algorithm is verified through navigation experiment with the single mobile robot, using the ROS-Gazebo simulator used for 3D simulation of multiple dynamic systems.

5. Impact of the DQN combined with GRU allowing action skipping is evaluated in comparison with the DQN combined with GRU not allowing action skipping and the DQN alone.

The rest of this paper is organized as follows. Section 2 presents the proposed algorithm. Section 3 describes the simulation environment and results. Section 4 concludes this paper.

## II. PROPOSED METHOD

Fig.2 shows the structure of the proposed method. The 4 consecutive(step) actions of mobile robot are taken as input of the DQN combined with the GRU.

As the output of the DQN combined with the GRU, Q-values of all the possible actions are obtained. Among the actions, the one making the largest Q-value is chosen as the new action when t mod Action Skipping is equal to 0. Otherwise, previous action is selected again. Proposed method aims at efficient navigation of mobile robot such as autonomous parking robot. Fig. 3 shows a real environment for robot navigation.

From the starting position, the robot tries to reach the goal point along the shortest path while avoiding obstacles. Our approach is to solve this task by the RL. In RL, the state consists of the current position of robot, the location of the obstacle, and the distance to the destination. The robot receives a positive reward when the distance to the destination is getting closer, and a negative reward is received if it collides with an obstacle or getting farther.

The robot is equipped with a LiDAR sensor for path planning.

### A. DQN with GRU

The structure of the DQN combined with the GRU is shown in Fig. 2. The combined structure show different first and the second layer as compared to the conventional DQN. The first layer is changed to Time Distributed Layer(TDL) playing the role for connecting its output to the GRU. The TDL connected to the GRU layer and two fully connected layers are implemented as parts of the DQN. Among the outputs of the last fully connected layer, the largest value is taken and selected as an action.

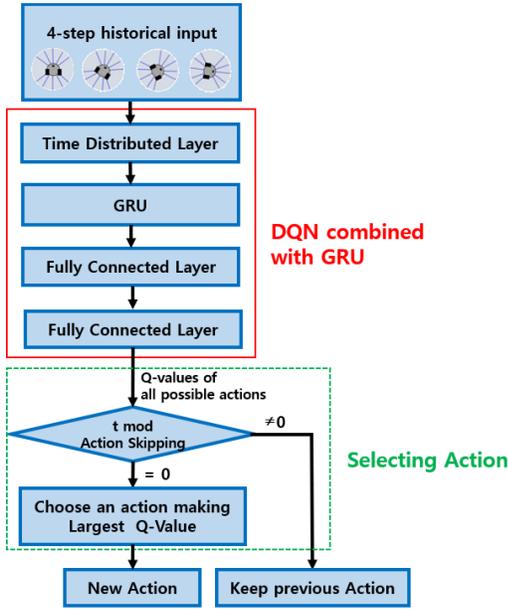

Fig. 2. Proposed method.

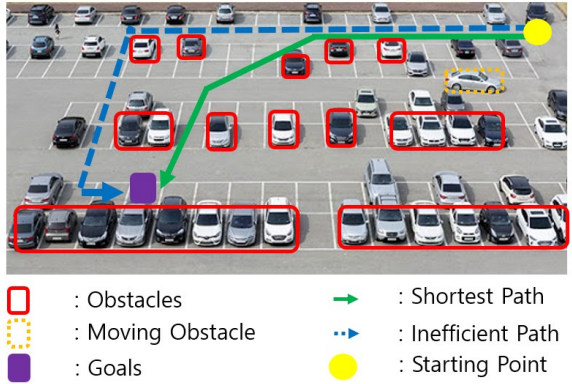

Fig. 3. Example of real environment – Parking lot.

For a policy $\pi$, Q value is given by

$$Q^{\pi}(s,a) := \mathbb{E}_{s,a,\pi}[\sum_{t=1}^{\infty} \gamma^{t} r_{t}] \quad (1)$$

where $\mathbb{E}$ is the expectation operator and $a$, $s$, $r$ are action, state, reward, respectively, and $\gamma$ is a discount factor and subscript/superscript $t$ represents time dependency. In words, the right hand side in (1) indicates the expected accumulated reward with time dependent discount.

The optimum Q value $Q^{*}(s,a)$ is defined as $\max_{\pi} Q^{\pi}(s,a)$. To deal with large dataset, a parameterized estimate of the Q-value function $Q(s,a;\theta)$, where $\theta$ is the weight parameter, is used.

The DQN is trained to minimize the loss function $L_i$ given by

$$L_i(\theta_i) = \mathbb{E}[(y_i - Q(s,a;\theta_i))^2] \quad (2)$$

where $i$ is the iteration index. The $Q(s,a;\theta_i)$ represents the Q value with the state $s$ and the action $a$ for the weight parameter $\theta_i$ in the $i$–th iteration.

where $i$ is the iteration index. The $Q(s,a;\theta_i)$ represents the Q value with the state $s$ and the action $a$ for the weight parameter $\theta_i$ in the $i$–th iteration. If the optimal value $Q^{*}(s',a')$ of the state $s'$ at the next time step is known for all the possible actions $a'$, then the optimal policy is to select the action maximizing the expected value. The target Q value $y_i$ in (2) is calculated by reward and next state as follows

$$y_i = r + \gamma \max_{a'} Q(s',a';\theta_{i-1}) \quad (3)$$

The parameter from the previous iteration $\theta_{i-1}$ is fixed when optimizing the loss function. During training, weights of the policy are updated in each iteration to minimize the loss function.

Through minimization of the difference between $y_i$ and $Q(s,a)$, the weight parameter $\theta_i$ is updated in next iteration as follows

$$\theta_{i+1} = \theta_i + \alpha \nabla_{\theta} L(\theta_i) \quad (4)$$

where $\alpha$ is the learning rate and $\nabla_{\theta}$ is the operator taking gradient with respect to $\theta$.

*B. DQN with action skipping*

In this work, mobile robot tries to reach the goal point along the shortest path while avoiding obstacles. It is necessary to select actions with high reward values for efficient learning process. The key-action defined as the most important action significantly contributing to the accumulated reward is critical for achieving the goal, i.e., reaching the goal point in this work [6]. However, if the chosen action for each step is changed everytime, it becomes difficult to identify which is the key-action. To solve this problem, action skipping is applied to the proposed algorithm. The Pseudo-code of the proposed algorithm can be seen in Table I where line 8 represents the action skipping. For simulations, the hyperparameter Actionskipping is set to 10. When $t$ mod Actionskipping = 0, newly chosen action is taken and otherwise previous action is kept.

III. SIMULATION ENVIRONMENT AND RESULTS

*A. Simulation environment*

Simulation is performed by the ROS-Gazebo simulator in a testbed to investigate the feasibility of solution obtained by the proposed method in real environment. When comparing with the navigation of mobile robot in real environment, the results obtained from the simulator tend to be more optimistic in the sense that real noise and occasional malfunction of mechanical components are not taken into account. The purpose of simulation is to show that the mobile robot is able to reach the dynamically varying goal points without making collisions. When the mobile robot gets closer to the goal point, it receives a positive reward, and when it gets farther it receives a negative reward. When the mobile robot arrives at the goal point, it gets a big positive reward and new goal point is randomly determined. Each episode ends when the mobile robot is collided with the boundary of the simulation environment or when 50sec of time is expired.

The testbed is shown in Fig. 4 (a) where red square box is a goal point. Agent, a mobile robot, is driven by two differential wheels as shown in Fig. 4 (b). The robot is equipped with a 360 degree 24-point LiDAR sensor which can measure distances and relative angles of the obstacles in the driving environment. The boundary walls in the testbed are also recognized as an obstacle. The initial position and goal point of the robot are given in absolute coordinates, and the position of the robot during simulation is calculated by odometry.

As shown in Fig. 5, the state parameters of the robot are defined as follows. The state is defined by heading angle of mobile robot, and current distance from goal point. heading angle of mobile robot, and current distance from goal point.

The reward of the robot is obtained through (4-7) as follows :

$$\theta = \frac{\pi}{2} + action \cdot \frac{\pi}{8} + \phi \quad (4)$$

$$R_\theta = 5 \cdot (1-\theta) \quad (5)$$

$$R_d = 2^{\frac{D_c}{D_g}} \quad (6)$$

$$R_{total} = R_\theta \cdot R_d \quad (7)$$

where $\theta$ is a heading angle of the robot, $\phi$ is yaw angle of mobile robot. The total reward defined as $R_{total}$ is obtained from angular reward $R_\theta$ and distance reward $R_d$. The $D_g$ is the initial distance to the goal point and the $D_c$ is the distance between current location of the robot and the goal point. The mobile robot is penalized by a big negative reward -100 when it is collided with the boundary. From the foregoing definitions, it is known that the angle error between the robot's heading angle and angle to the goal point is smaller and the total reward is larger when the robot becomes closer to the goal point. Since the robot has constant linear velocity 0.15 m/s, the actions of the robot are learned through five actions corresponding to five different angular velocities listed in Table II. Table III shows the training parameters and values of the proposed method during the simulation.

TABLE I.  PSEUDO-CODE OF DQN COMBINED WITH GRU ALLOWING ACTION SKIPPING.

| | Algorithm 1. DQN with Action skipping. |
|---|---|
| 1 | Initialize replay memory $D$ to capacity $N$ |
| 2 | Initialize action-value fucntion $Q$ with random weights |
| 3 | **for** episode =1, $M$ **do** |
| 4 | Initialize sequence $s_1 = \{x_1\}$ and preprocessed sequenced $\phi_1 = \phi\{s_1\}$ |
| 5 | **for** $t$ =1, $T$ **do** |
| 6 | With probability $\varepsilon$ select a random action $a_t$ |
| 7 | otherwise select $a_t$ where, |
| 8 | $a_t = \begin{cases} a_{t-1} & \text{if } t \bmod \text{Actionskipping} \neq 0 \\ \max_a Q^*(\phi(s_t),a;\theta) & \text{if } t \bmod \text{Actionskipping} = 0 \end{cases}$ |
| 9 | Excute action $a_t$ in emulator and observe reward $r_t$ |
| 10 | Set and preprocess $\phi_{t+1} = \phi(s_{t+1})$ |
| 11 | Store transition $(\phi_t, a_t, r_t, \phi_{t+1})$ in $D$ |
| 12 | Sample random Mini-batch or transitions $(\phi_j, a_j, r_j, \phi_{j+1})$ from D |
| 13 | Set $y_j = \begin{cases} r_j & \text{for terminal } \phi_{j+1} \\ r_j + \gamma \max_{a'} Q^*(\phi(s_t),a';\theta) & \text{for non-terminal } \phi_{j+1} \end{cases}$ |
| 14 | Perform a gradient descent step on $(y_j - Q(\phi_j, a_j;\theta))^2$ |
| 15 | **end for** |
| 16 | **end for** |

*B. Simulation results*

In this subsection, simulation results are presented. The simulation results are obtained after 500, 1000, and 3000 training episodes. The plot in Fig. 6 (a) is the time-varying total reward obtained by DQN alone. It can be seen that the total reward tends to diverge due to occasional collisions of the robot with the boundary. Fig. 6 (b) is the variation of the total reward obtained by the DQN with GRU allowing action skipping. As seen in the Figure, the total reward tends to less diverge, representing that the frequency of collision with the boundary is reduced as compared to the case of DQN alone. Fig. 7 shows the variation of average Q-value. As shown in Fig. 7, the Q-value of the proposed algorithm is increased faster that the DQN alone. Note that the scaling of the two figures in Fig.7 is different, i.e., Fig. 7 (a) shows negative Q-values along each episode.

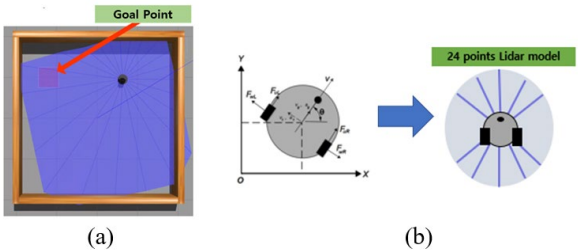

(a)　　　　　　　(b)

Fig. 4.　Simulation testbed

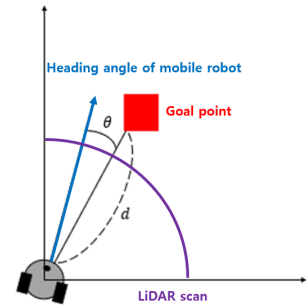

Fig. 5.　Graphical definition of state.

TABLE II.    5 DISCRETE ACTION FOR ROBOT.

|  | Angular velocity (rad/s) |
|---|---|
| Action 1 | -1.5 |
| Action 2 | -0.75 |
| Action 3 | 0 |
| Action 4 | 0.75 |
| Action 5 | -1.5 |

TABLE III.    TRAINING PARAMETERS AND THEIR VALUES.

| *Parameter* | *Value* |
|---|---|
| Batch size | 64 |
| Target update rate | 2000 |
| Discount factor | 0.99 |
| Learning rate | 0.75 |
| Epsilon (probability) | 1.5 |
| Epsilon decay | 0.99 |
| Epsilon min | 0.05 |

Table IV summarizes the results of the all the methods after 500, 1000, and 3000 episodes. Success rate is defined as the portion of evaluation episodes ended in reaching goal point without experiencing collision with the boundary. In the case of 500 episodes, there are many cases of failure because of the collision with the boundary. After 1000 episodes of training with proposed algorithm, the time taken to reach the goal point is decreased to 11.2seconds, and the success rate is increased to 65%. After the 3000 episodes, the time taken to reach the goal point is further decreased to 7.3 seconds and the success rate is improved to 87%. It can be confirmed that the overall performance of the policy learned by the DQN combined with GRU allowing action skipping is the best among three algorithms considered.

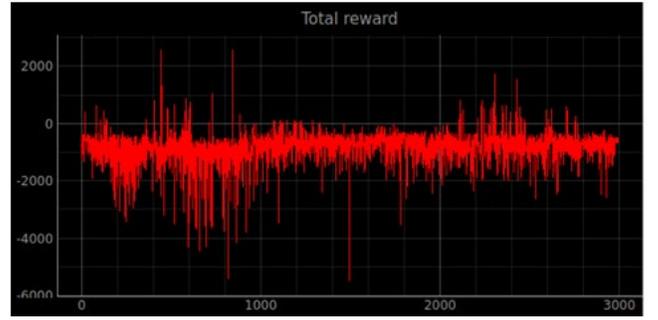

(a)

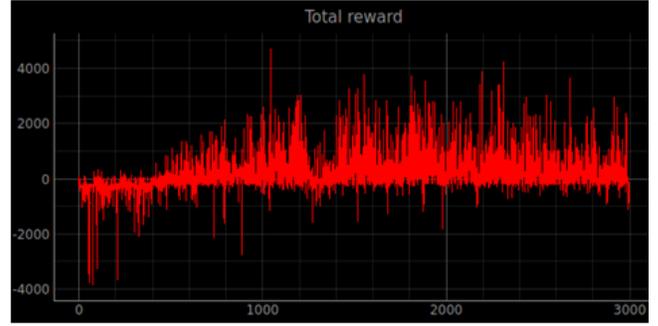

(b)

Fig. 6. Variation of total reward according to number of episodes: (a) DQN alone; (b) DQN combined with GRU allowing action skipping (proposed algorithm)

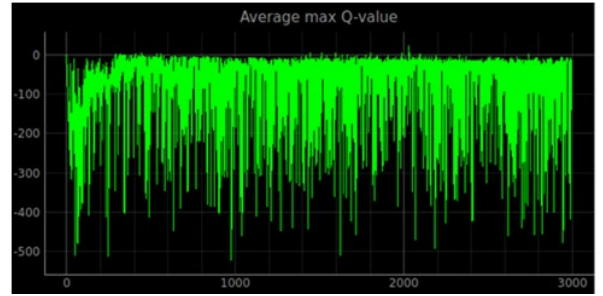

(a)

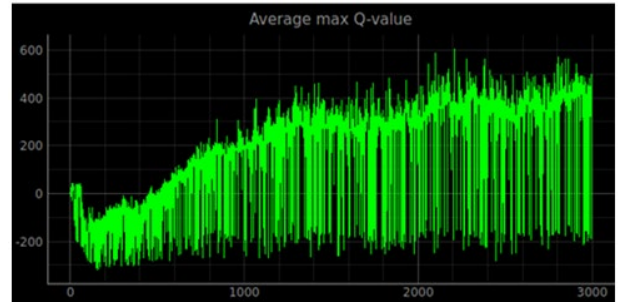

(b)

Fig. 7. Variation of max Q-value according to number of episodes: (a) DQN alone; (b) DQN combined with GRU allowing action skipping (proposed algorithm)

TABLE IV. AVERAGE TIME AND SUCCESS RATE OF MOBILE ROBOT TO GOAL POINT

|  | 500 episodes | 1000 episodes | 3000 episodes |
|---|---|---|---|
| Average time to reach goal point (DQN + GRU + action skipping) | 13.5 sec | 11.2 sec | 7.3 sec |
| Success rate (DQN + GRU + action skipping) | 48% | 65% | 87% |
| Average time to reach goal point (DQN+GRU) | 15.2 sec | 12.6 sec | 10.8 sec |
| Success rate (DQN+GRU) | 30% | 49% | 57% |
| Average time to reach goal point (DQN alone) | 15.7 sec | 13.9 sec | 12.8 sec |
| Success rate (DQN alone) | 26% | 32% | 46% |

## IV. CONCLUSION

In this paper, we propose a path planning algorithm for mobile robot by using DQN with GRU allowing action skipping. Path planning is performed for each goal point. Performance comparison of the proposed algorithm with DQN alone and DQN combined with GRU is made with the ROS-Gazebo simulator for a testbed. By allowing action skipping, the proposed algorithm demonstrates superior performances measured by average time taken to reach goal point and success rate. Through action skipping, the policy is able to learn key-action more efficiently. As a result, the policy trained with the proposed algorithm is seen to be most efficient in navigation of mobile robot.